\documentclass[11pt]{article}
\usepackage[preprint]{acl}
\usepackage{times}
\usepackage{latexsym}
\usepackage[T1]{fontenc}
\usepackage[utf8]{inputenc}
\usepackage{microtype}
\usepackage{inconsolata}
\usepackage{booktabs}
\usepackage{amsmath}
\usepackage{url}
\usepackage{graphicx}
\usepackage{xcolor}

\usepackage{pdfpages}

\title{Mixed Feelings, Single Polarity}

\author{Emily \"{O}hman \\
  Waseda University, Japan \\
  \texttt{ohman@waseda.jp} \\\And
  Saif Mohammad \\
  National Research Council Canada \\
  \texttt{saif.mohammad@nrc-cnrc.gc.ca} \\}

\begin{document}
\maketitle

\begin{abstract}

Emotion datasets are routinely mapped onto positive and negative valence, by assigning each emotion category one polarity. We test where this works and where it fails, using SemEval-2018 Affect in Tweets (English, Arabic, Spanish), Affective Text aligned with EmoBank, raw GoEmotions annotations, naturalistic hashtag corpora, and a controlled annotation study. Fixed projection works well when a text carries labels of one valence. It fails on a subset that can be identified from the labels alone: tweets whose emotion labels point in opposite directions. These are 9.0\% of polar SemEval test tweets but cause 54.8\% of the polarity errors of a learned projection, are more than twice as likely to receive a neutral-or-mixed gold label, and are not rescued by modeling the full label set. Annotators produce such configurations routinely: 16.3\% of multilabel GoEmotions judgments combine emotions from the dataset's own positive and negative parents. In a controlled study holding word and reader constant, context changes the coarse valence judgment in 56.8\% of cases, against 19.9\% for controls. We recommend projecting label configurations rather than labels, abstaining or keeping a mixed class where labels conflict, and reporting coverage alongside accuracy.
\end{abstract}

\section{Introduction}

Suppose an emotion classifier assigns a tweet both \textit{joy} and \textit{sadness}, and a downstream system needs a single polarity. A category-to-valence lookup gives opposite signs to the two labels, leaving the tweet-level decision unresolved. In GoEmotions \citep{demszky-etal-2020-goemotions}, 16.28\% of multilabel emotional judgments combine categories from the dataset's positive and negative parents. SemEval-2018 Affect in Tweets \citep{mohammad-etal-2018-semeval} provides independently annotated text-level valence with which to assess what those labels predict. Fixed mappings can provide useful category-level priors without determining the polarity of every text.

Mappings of this kind are widespread. Emotion datasets use small basic-emotion sets \citep{ekman1992argument,plutchik1980emotion}, larger multilabel inventories such as GoEmotions \citep{demszky-etal-2020-goemotions}, or continuous dimensions such as valence, arousal, and dominance (VAD) \citep{russell1980circumplex,buechel-hahn-2017-emobank}. Combining or comparing them requires a mapping between schemes, often onto coarse valence. Corpora are harmonized through such mappings \citep{bostan-klinger-2018-analysis,koufakou2024towards,el-haj-takanami-2023-unifying}, and inventories are coarsened through them \citep{gong-etal-2024-mapping}. The relevant claim is that an emotion category can be assigned a fixed valence across the items to which it is applied. This assumption is often implemented during preprocessing without being separately validated against item-level valence judgments. When it fails, distinct labels merge, predictions that were wrong become correct after projection, and information about intensity, context, and multilabel composition disappears. A classifier that predicts \textit{anger} for a text annotated \textit{disgust}, for instance, is wrong at the fine-grained level but correct once both are scored as negative.

Let $x$ denote the text unit being annotated (e.g., a word, sentence, or tweet), and let $E(x)=e$ be an emotion label assigned to it by a human or model. A fixed collapse makes a simple claim: knowing $e$ is sufficient to assign $x$ a fixed valence,
\begin{equation}
    E(x)=e \Longrightarrow V(x)=f(e).
    \label{eq:mapping}
\end{equation}
An uncertainty-aware alternative instead estimates $P(V(x)\mid E(x)=e)$, allowing units with the same emotion label to differ in valence. We test the adequacy of the fixed projection, not whether emotions have characteristic valence. For example, \textit{I'm furious about the delay, but thrilled we made it home} may carry anger while being positive overall. This is a failure of label-to-text projection, not evidence that anger itself is positive.

Study~1 examines the information emotion labels provide about valence, how that information changes with co-occurring labels, and how the patterns extend to naturalistic emotion signals. We analyze SemEval-2018 \citep{mohammad-etal-2018-semeval}, Affective Text headlines \citep{strapparava-mihalcea-2007-semeval} aligned with EmoBank \citep{buechel-hahn-2017-emobank}, raw GoEmotions annotations \citep{demszky-etal-2020-goemotions}, and two naturalistic hashtag corpora \citep{mohammad-kiritchenko-2015,vishnubhotla-mohammad-2022-tusc}. A fixed lexical projection assumes that an affective prior associated with a word or sense remains informative when that item occurs in context. Study~2 tests this assumption by comparing affective judgments of words in isolation with judgments of the same words in sentence contexts. Together they address three questions:
\begin{description}
    \item[RQ1:] How well does an assigned emotion label predict the valence of the same annotated text unit, and is this relation stable across languages, datasets, and annotation designs?
    \item[RQ2:] How does text-level valence vary with co-occurring emotion labels, where does label-only polarity projection fail, and how much information is retained under the GoEmotions sentiment hierarchy?
    \item[RQ3:] With lexical item and reader held constant, how much does context change coarse valence, dominant emotion, and position in affective space?
\end{description}

Across studies, category prototypes are informative but do not reliably determine text-level valence, while context alters judgments of the same word. Table~\ref{tab:recommendation} summarizes our projection recommendations.\footnote{Code: \url{https://anonymous.4open.science/r/VAD_collapse-BE76/}}

\section{Related Work}

Categorical and dimensional representations capture related but non-identical affective structure \citep{ekman1992argument,plutchik1980emotion,russell1980circumplex}, and each can be recovered from the other with high accuracy: \citet{buechel-hahn-2018-emotion} map between VAD and basic-emotion ratings with near-human reliability, including cross-lingually, and \citet{park-etal-2021-dimensional} predict VAD scores from corpora annotated only with categories. Recovering one representation from another on average does not mean that a single category fixes the valence of every text carrying it. Harmonization nonetheless makes this step routinely: \citet{el-haj-takanami-2023-unifying} average the VAD vectors of a sentence's emotions into one target and threshold predicted valence into three classes, and \citet{gong-etal-2024-mapping} map fine-grained inventories down to sentiment classes. Reviews caution that unified data must be read in light of differing emotion definitions and annotation procedures \citep{koufakou-nieves-2025-review}, and that lexicon-derived affect scores are easily over-interpreted \citep{mohammad2016practical}.

Psychology gives reasons to expect variation within categories. Positive and negative affect can co-occur in experience, as in happiness and sadness, fear and enjoyment, or disgust and amusement \citep{larsen-mcgraw-2014}. This has led to bivariate measures that rate positive and negative affect separately instead of asking for one net valence \citep{kreibig-gross-2017}. If people can feel both at once, a coarse representation of a text may need to keep both as well. Emotion categories are connected by continuous gradients rather than separated by clean boundaries \citep{cowen-keltner-2017-selfreport}, appraisal approaches derive emotion from context-sensitive evaluations of events \citep{hofmann-etal-2020-appraisal,troiano-etal-2023-dimensional}, and distributed label representations capture graded relations among categories \citep{wang-zong-2021-distributed}. Lexical resources show the same: the NRC Emotion Lexicon pairs Plutchik categories with sentiment \citep{mohammad-turney-2010-emotions,mohammad2013nrc}, yet many entries depend on word sense and context \citep{zad-etal-2021-hell}, and classifiers degrade sharply once explicit lexical cues are removed \citep{sosea-caragea-2025-hard}.

Annotation practice points the same way. \citet{alm-etal-2005-emotions} split surprise into positive and negative variants, multilabel resources such as XED let emotions co-occur within a text \citep{ohman-etal-2020-xed}, and category overlap is a recurring annotation problem \citep{ohman2020rethinking}. GoEmotions makes the mapping explicit: its 27 categories come with a sentiment hierarchy based on \citet{cowen-keltner-2017-selfreport}, and collapsing to sentiment or Ekman groupings improves classification \citep{demszky-etal-2020-goemotions}. That work does not claim that every judgment falls under a single parent, so we treat the hierarchy as a strong example of a useful coarse representation and ask how closely raw annotations respect it.

Concurrent work asks whether same-named emotion labels are operationally equivalent across datasets \citep{anonymous-concurrent}; we instead ask whether an emotion label determines the valence of the text it annotates.

\section{Study 1: Corpus evidence}

Study~1 evaluates emotion-to-valence inference using independently annotated valence, co-occurring emotion labels, within-annotator judgments, and naturalistic hashtag signals. Bootstrap, null-model, and screening details are given in Appendix~\ref{app:methods}.

\subsection{How informative are emotion labels about text valence?}
\label{sec:study1a}

As part of SemEval-2018 Task~1, \citet{mohammad-etal-2018-semeval} obtained separate annotations for categorical emotion and continuous valence. The emotion-classification dataset (E-c) uses eleven labels (anger, anticipation, disgust, fear, joy, love, optimism, pessimism, sadness, surprise, trust), while valence regression (V-reg) uses Best--Worst Scaling \citep{kiritchenko2017best} to produce continuous scores on a 0--1 scale. Neither annotation is derived from the other. Matching items present in both datasets by identifier yields 6,563 tweets (2,355 English, 1,794 Arabic, 2,414 Spanish), retaining the original train/dev/test splits.

For the SemEval-2018 analysis, emotion information comes from E-c and valence from the independently annotated V-reg data. For each emotion $e$, tweets annotated with $e$ form one group and all remaining tweets the other; continuous valence is used to rank the two groups. We retain both the direction and strength of this association:
\begin{equation}
    S_e = 2(\mathrm{AUC}_e-0.5),
    \qquad
    D_e = |S_e|.
\end{equation}
Thus $S_e>0$ means tweets carrying $e$ tend to receive higher valence scores, whereas $S_e<0$ means they tend to receive lower scores; $D=0$ indicates no ranking information and $D=1$ perfect separation. Both measures are threshold-free and insensitive to class prevalence. We estimate each emotion's continuous prototype as its mean training-set V-reg score. For polarity analyses, we use the official SemEval ordinal valence labels (V-oc), derived from V-reg using language-specific boundaries: classes $-3$ to $-1$ are negative, $0$ is neutral or mixed, and $+1$ to $+3$ are positive. The value $0.5$ in the AUC transformation denotes chance-level discrimination, not a neutral boundary on the V-reg scale.

Because E-c is multilabel, the primary diagnosticity and prototype analyses treat a tweet as carrying emotion $e$ whether or not other emotion labels are also present. We use single-label tweets for a separate projection test, where each observed valence score can be compared with exactly one train-estimated emotion mean. On these held-out tweets, the emotion mean predicts valence far better than a global mean ($R^2=.734$ [95\% CI $.664,.779$] English, $.583$ [$.418,.696$] Arabic, $.628$ [$.554,.685$] Spanish). Emotion-level mean valence is also similarly ranked across languages (test-set $\rho=.873$ English--Arabic, $.991$ English--Spanish, $.882$ Arabic--Spanish), supporting stable category-level valence prototypes.

Joy is almost perfectly valence-diagnostic ($D=.879$), and love, sadness, anger, and optimism are strong on average (Table~\ref{tab:semeval}). Anticipation ($D=.107$) and surprise ($D=.076$) are weak. Anticipation's signed association with continuous valence differs between Arabic and Spanish, but under the official V-oc classes, negative judgments outnumber positive ones in all three languages. Thus differences in continuous association do not establish opposite categorical polarity. Prototype strength varies considerably by emotion category.

\begin{table}[t]
\centering
\small
\begin{tabular}{ll}
\toprule
Emotion & $D$ [95\% CI] \\
\midrule
Joy          & .879 [.859,.898] \\
Love         & .713 [.667,.754] \\
Sadness      & .679 [.639,.716] \\
Anger        & .646 [.611,.681] \\
Optimism     & .644 [.599,.687] \\
Disgust      & .591 [.551,.631] \\
Pessimism    & .514 [.465,.560] \\
Fear         & .445 [.383,.505] \\
Trust        & .345 [.269,.426] \\
Anticipation & .107 [.057,.172] \\
Surprise     & .076 [.029,.171] \\
\bottomrule
\end{tabular}
\caption{Threshold-free valence diagnosticity in SemEval-2018, averaged across English, Arabic, and Spanish test sets, with 95\% bootstrap intervals. Per-language estimates appear in Appendix Table~\ref{tab:perlangfull}.}
\label{tab:semeval}
\end{table}

\paragraph{Affective Text $\times$ EmoBank.}
\label{sec:study1b}

To check that the pattern is not specific to SemEval or to tweets, we align SemEval-2007 Affective Text headlines \citep{strapparava-mihalcea-2007-semeval}, which carry continuous intensity scores for six emotions, with EmoBank valence \citep{buechel-hahn-2017-emobank}, giving 1,192 headlines. Following the original Affective Text criterion, an emotion counts as present at intensity $\geq50$.

The ordering replicates (Table~\ref{tab:emobank}). Joy ($D=.843$) and sadness ($D=.832$) are highly diagnostic; disgust ($.661$), fear ($.609$), and anger ($.603$) are intermediate; surprise is weak ($D=.220$ [.065,.370]). Headlines with surprise average 3.015 on EmoBank's 1--5 valence scale, essentially the midpoint. The ordering holds at a $\geq25$ threshold, with continuous intensity instead of a threshold, and for reader- and writer-perspective valence separately (Appendix Table~\ref{tab:emobankpersp}).

\begin{table}[t]
\centering
\scriptsize
\begin{tabular}{lrlrl}
\toprule
Emotion & $n$ & $D$ [95\% CI] & Mean $V$ & $\rho$ [95\% CI] \\
\midrule
Joy      & 144 & .843 [.808,.875] & 3.376 & +.773 [.749,.795] \\
Sadness  & 139 & .832 [.787,.873] & 2.260 & -.709 [-.737,-.679] \\
Disgust  & 19  & .661 [.539,.769] & 2.383 & -.407 [-.454,-.357] \\
Fear     & 114 & .609 [.535,.678] & 2.463 & -.562 [-.602,-.521] \\
Anger    & 38  & .603 [.493,.706] & 2.421 & -.558 [-.596,-.518] \\
Surprise & 49  & .220 [.065,.370] & 3.015 & +.143 [.086,.197] \\
\bottomrule
\end{tabular}
\caption{Affective Text $\times$ EmoBank, combined valence. $D$ uses the $\geq50$ presence threshold; $\rho$ uses continuous intensity. At $\geq25$ the ordering is unchanged (joy .820, sadness .735, surprise .182; surprise mean valence 2.958).}
\label{tab:emobank}
\end{table}

\subsection{What happens when emotion labels co-occur?}

\paragraph{Co-occurrence (SemEval-2018).}
For each ordered emotion pair $(e,c)$, we compute
\[
\Delta V_{e,c}
=
\overline{V}(e,c)-\overline{V}(e,\neg c),
\]
the difference between the mean valence of tweets carrying both $e$ and $c$ and tweets carrying $e$ without $c$. Thus $\Delta V>0$ means that tweets carrying $e$ have higher valence when $c$ is also present, and $\Delta V<0$ means lower valence. Of the 110 possible ordered pairs, 68 had at least five $e+c$ tweets in each language in the training data; 62 had the same sign of $\Delta V$ in English, Arabic, and Spanish. We tested these directional effects on held-out data: 46 retained the same sign in all three languages, and 45 had pooled 95\% intervals excluding zero (Appendix Tables~\ref{tab:cooc-full1}--\ref{tab:cooc-full3}). For example, anger tweets have mean valence $.202$ higher when joy is also annotated, whereas joy tweets have mean valence $.199$ lower when anger is also annotated. These are tweet-level associations and do not identify which span or emotion contributes the valence difference.


\paragraph{What polarity projection loses.}
We evaluate label-only polarity projection on 2,076 held-out SemEval tweets with at least one emotion label, using official V-oc polarity as the target. A conventional lookup assigning nine emotions fixed signs and leaving anticipation and surprise unsigned covers 92.1\% of tweets, with 20.0\% errors among predictions across the three gold classes. Restricting evaluation to gold-positive and gold-negative tweets, its error rate falls to 3.5\% at 94.3\% coverage. Fixed projection is therefore highly effective when the target has a clear polarity.

Errors are concentrated in particular label configurations. Tweets containing both conventionally positive- and negative-associated labels constitute 11.8\% of the labeled test set, but 38.2\% have neutral-or-mixed V-oc labels, compared with 16.4\% of other tweets. A model averaging label-specific polarity priors makes errors on 33.6\% of these opposed-label tweets with polar gold labels, against 2.7\% of other polar tweets; among tweets with polar gold labels, opposed configurations constitute 9.0\% of cases but account for 54.8\% of the model's wrong-polarity predictions. Estimating priors for exact label sets gives nearly identical performance. These results identify where the tested label-only projections are unreliable, including cases whose gold polarity is unambiguously positive or negative (Table~\ref{tab:projection}).

\begin{table*}[t]
\centering
\small
\setlength{\tabcolsep}{4pt}
\begin{tabular}{lrrrrrrr}
\toprule
& \multicolumn{2}{c}{All ($n=2076$)}
& \multicolumn{2}{c}{Polar ($n=1681$)}
& \multicolumn{2}{c}{Opposed ($n=246$)}
& \\
\cmidrule(lr){2-3}
\cmidrule(lr){4-5}
\cmidrule(lr){6-7}
Method & Cov. & Err. & Cov. & Err. & Cov. & Err.
& Pred.\ 0 \\
\midrule
Fixed lookup, unsigned & 92.1 & 20.0 & 94.3 & 3.5
& 48.8 & 55.0 & 0 \\
Fixed lookup, positive & 92.0 & 21.4 & 93.5 & 4.5
& 55.7 & 57.7 & 0 \\
Averaged priors, binary & 100 & 23.5 & 100 & 5.5
& 100 & 58.9 & 0 \\
Label-set priors, binary & 100 & 23.6 & 100 & 5.6
& 100 & 59.4 & 0 \\
Selective projection & 88.2 & 18.5 & 91.4 & 2.8
& 15.4 & 55.3 & 0 \\
Averaged priors, 3-class & 100 & 23.0 & 100 & 6.0
& 100 & 59.4 & 30 \\
Label-set priors, 3-class & 100 & 22.9 & 100 & 8.2
& 100 & 57.7 & 114 \\
\bottomrule
\end{tabular}
\caption{Held-out SemEval-2018 emotion-to-valence projection using official V-oc labels. Cov.\ is prediction coverage (\%); Err.\ is the percentage of incorrect predictions among covered items. All includes negative, neutral-or-mixed, and positive gold classes. Polar excludes gold class 0; Opposed contains conventional positive- and negative-label combinations and includes all three gold classes. Pred.\ 0 is the number of neutral-or-mixed predictions across the full test set. The selective projection is an illustrative abstention baseline. Polar-only evaluation is a retrospective diagnostic subset, not a deployable selection procedure.}
\label{tab:projection}
\end{table*}

\paragraph{GoEmotions.}
\label{sec:study1c}

GoEmotions assigns 12 of its 27 emotions to a positive parent, 11 to a negative parent, and four (realization, surprise, curiosity, confusion) to an ambiguous parent, with neutral kept separate \citep{demszky-etal-2020-goemotions}. We use the raw release, which preserves each annotator's complete multilabel response: 211,225 rows, of which we drop 3,411 marked \texttt{example\_very\_unclear} (none has a selected label), leaving 207,814 judgments over 58,009 Reddit comments. We map each selected emotion to its published parent and ask whether a single judgment contains both a positive- and a negative-parent emotion. Since the hierarchy is the dataset's own, this is a test of GoEmotions against its own annotators.

\paragraph{Annotators respect the hierarchy.}
Of 35,994 multilabel emotional judgments, 16.28\% combine positive- and negative-parent emotions (95\% rater-cluster CI [14.68, 18.26]). A matched null that preserves label cardinality and label frequencies predicts 39.53\%, and none of 10,000 null datasets produces a rate as low as observed (Table~\ref{tab:goemotions}). Every one of the 27 categories is more parent-consistent than its own matched null (Appendix Table~\ref{tab:goemotioncats}).

\paragraph{But one multilabel judgment in six crosses it.}
Positive--negative co-selection rises with length, from 1.08\% of emotional judgments in the shortest token-length quintile to 6.04\% in the longest, yet among the 18,713 multilabel judgments on single-sentence comments the rate is still 15.14\% (null 41.02\%). Strongly valenced categories cross too: 33.0\% of multilabel judgments containing sadness also contain a positive emotion, and 37.7\% of those containing remorse. Consensus filtering with the published two-annotator rule leaves 2.68\% of non-neutral item label sets with both parents. The result is stable when the 1--10 most prolific raters are removed (3.51--3.93\% of all emotional judgments, against 3.84\% overall; Table~\ref{tab:raterdrop}).

\begin{table}[t]
\centering
\scriptsize
\begin{tabular}{lrr}
\toprule
Measure / denominator & Observed & Null \\
\midrule
P+N, multilabel judgments & 16.28\% & 39.53\% \\
P+N, one-sentence multilabel & 15.14\% & 41.02\% \\
Any-parent, multilabel judgments & 39.40\% & 66.90\% \\
P+N, all emotional judgments & 3.84\% & 9.31\% \\
Any-parent, all emotional judgments & 9.30\% & 15.83\% \\
P+N, filtered item label sets & 2.68\% & --- \\
\midrule
Coarse entropy $H(C)$ & 2.23 / 3 bits & --- \\
Residual $H(F\mid C)$ & 3.11 bits & --- \\
\bottomrule
\end{tabular}
\caption{GoEmotions judgments crossing the published sentiment hierarchy. P+N: both positive- and negative-parent emotions selected. Each observed rate is compared with a null over the same denominator (Appendix~\ref{app:methods}).}
\label{tab:goemotions}
\end{table}

\subsection{Do the patterns extend to naturalistic emotion signals?}
\label{sec:study1d}

Self-labeled hashtags give an independent naturalistic signal that no annotator assigned. We use the Tweet Emotion Corpus (TEC; \citealp{mohammad-kiritchenko-2015}), which contains 21,051 tweets collected with emotion-word hashtags. After removing hashtags, URLs, and mentions, lexical valence separates joy from the pooled negative categories only moderately ($D=.333$). As a second check, we analyze TUSC social-media posts \citep{vishnubhotla-mohammad-2022-tusc} as distributed through ABCDE \citep{wahle-etal-2026-affect}. The strict set yields 332 analyzable posts (97 joy, 235 pooled negative). Residual lexical valence distinguishes joy from the pooled negative categories with AUC $=.716$ [95\% CI $.654,.776$], $D=.433$ [$.308,.552$]. Among the 231 strict-set posts containing at least one extreme lexical cue ($|V|\geq.70$), 26.8\% contain extreme cues exclusively opposite to the hashtag-derived valence. Full preprocessing details, per-emotion comparisons, and robustness checks are reported in Appendix~\ref{app:hashtags}.


\begin{figure*}[t]
\centering
\includegraphics[width=\linewidth]{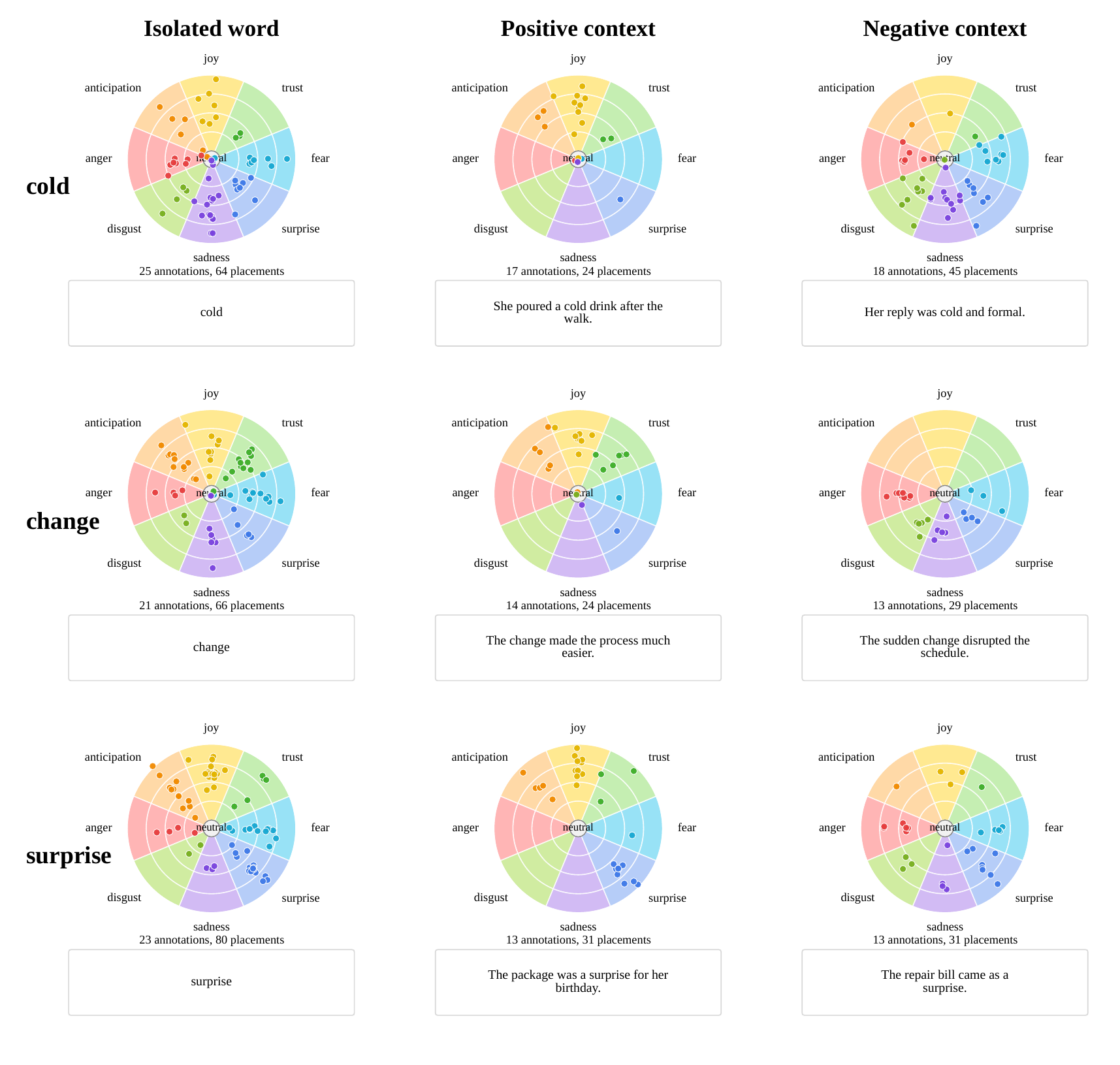}
\caption{Observed wheel placements for three illustrative lexical targets in isolation and in their two contextualized uses. Each dot is an observed Study~2 main-arm association point. The contextual columns are ordered by their observed coarse-valence balance for display only; all inferential analyses use the complete set of 50 targets.}
\label{fig:study2examples}
\end{figure*}
\section{Study 2: Same word, different context}
\label{sec:study2}

Study~2 tests whether context changes a reader's valence and emotion judgments of the same word, comparing isolated and contextualized uses within participants.

\subsection{Design}

\paragraph{Materials.}
We designed contextual contrasts to examine how affective judgments vary with lexical meaning, evaluative framing, mixed affect, event-dependent affect, and register. Sense-disambiguation items contrasted different meanings of a word; evaluative reframing changed how a referent or situation was appraised; mixed-affect items combined positive and negative associations; event-dependent items varied the surrounding situation; and register items contrasted conventional and colloquial usage. Stability controls were designed to preserve the expected affective reading. Isolated trials presented the target word without sense information or sentence context; contextualized trials presented the same word within a sentence. We selected 50 diagnostic, non-representative English targets using the NRC Emotion, Emotion Intensity, and VAD lexicons \citep{mohammad-turney-2010-emotions,mohammad2013nrc,mohammad-etal-2018-semeval,vad-acl-2018} and earlier lexical disagreement patterns (Appendix Table~\ref{tab:legacy-lexical}). The set comprised 13 sense-disambiguation, 17 evaluative-reframing, eight mixed-affect, six event-dependent, two register, and four control targets, yielding 50 isolated-word and 100 contextualized stimuli.

\paragraph{Instrument and procedure.}
Annotators used an extended version of Sentimentator \citep{ohman-etal-2018-creating} with an inverted Plutchik-style wheel: angle encodes emotion category and distance from the center encodes intensity. They placed one to eight points on the wheel and separately chose Positive, Negative, Both, or Neither. In the main arm, each participant was scheduled to see every target in isolation first and in its two contexts later, with at least eight intervening trials. Completion of all 150 trials was not required. We compare each contextualized judgment with the same participant's earlier judgment of that word in isolation, including only cases where both judgments are available. Exploratory arms that presented context before isolation were excluded because prior exposure to the sentence could influence the subsequent isolated-word judgment.

\paragraph{Measures.}
For an annotation with $k$ points $\mathbf{p}_1,\ldots,\mathbf{p}_k$, normalized by the wheel radius, we compute the centroid $\boldsymbol{\mu}$, mean intensity $I$, directional coherence $C$, and spread $S$:
\begin{equation}
\begin{gathered}
\boldsymbol{\mu}=\tfrac{1}{k}\textstyle\sum_i\mathbf{p}_i, \quad
I=\tfrac{1}{k}\textstyle\sum_i\lVert\mathbf{p}_i\rVert, \\
C=\lVert\boldsymbol{\mu}\rVert/I, \quad
S=\sqrt{\tfrac{1}{k}\textstyle\sum_i\lVert\mathbf{p}_i-\boldsymbol{\mu}\rVert^2}.
\end{gathered}
\end{equation}
Several strong points in opposite directions give high $I$ but low $C$. The main context effect is displacement $\Delta_E=\lVert\boldsymbol{\mu}_{\mathrm{context}}-\boldsymbol{\mu}_{\mathrm{isolated}}\rVert$ within participant and target. Because a centroid can hide a change in the dominant emotion, we also track the \emph{strongest association}, the selected point with the greatest radial intensity, and map it to one of the wheel's eight 45-degree basic-emotion sectors, or to Neutral if $r<.10$. We report pair-weighted summaries and participant-equal bootstrap estimates, since completion varied. Mechanism comparisons use stability controls as the reference, with two-way cluster-robust standard errors by participant and target and standardized trial position as a covariate.

\subsection{Results}

Of 35 participants assigned to the main arm, 29 contributed at least one valid isolated--context pair. The analysis covers 2,810 annotations and 7,373 wheel points across all 50 targets (8--29 annotations per stimulus, median 17), giving 1,617 within-participant pairs. In every pair the isolated judgment came first.

\paragraph{Same reader, same word, different valence.}
When participants saw a target again in context, their Positive/Negative/Both/Neither judgment changed in 56.8\% of pairs (participant-equal 57.8\%, 95\% CI [54.2, 61.4]). Some change is ordinary response variability: stability controls, whose contexts were written to keep the isolated reading, switch in 19.9\% of pairs. Targets whose contexts were designed to change the reading switch at 2.2 to 3.5 times that rate (44.5--68.8\%; Table~\ref{tab:study2mechanisms}).

\paragraph{Mixed readings resolve.}
In isolation, 23.2\% of judgments were \textit{Both}; in context, 8.9\% were. Only 13.9\% of isolated-\textit{Both} judgments stayed \textit{Both}. The rest went almost evenly to Positive (39.2\%) and Negative (34.9\%). A word that a reader holds as ambivalent on its own is usually committed to one side in context, and the context decides which side.

\paragraph{The dominant emotion changes too.}
The sector of the strongest association changed in 70.4\% of pairs; moving the Neutral cutoff between $.08$ and $.12$ changes this by at most 0.1 points. This measure is noisier than the coarse label: even stability controls switch sector in 55.3\% of pairs, against 72.0\% for manipulated targets. The strongest association rotated by $71.9^\circ$ on average (median $56.2^\circ$) and moved .673 wheel units (participant-equal .711), while its intensity barely changed ($\Delta r=-.021$). Among the 925 switches from one non-neutral emotion sector to another, 63.8\% moved to a non-adjacent sector, so the overall switch rate is not driven mainly by points crossing a neighboring sector boundary. Figure~\ref{fig:study2transitions} shows the full transition matrix.

\paragraph{Context does not shift intensity.}
Centroids moved by $\Delta_E=.443$ on average (participant-equal $.409$, [.372, .444]), about 40\% of the wheel radius. Mean intensity did not change ($\Delta I=.000$, [$-.023$, .026]). Multi-point responses instead became more coherent ($\Delta C=.069$, [.047, .093]) and less spread out ($\Delta S=-.052$, [$-.069$, $-.034$]): context pulled a reader's associations toward one direction. Figure~\ref{fig:study2examples} shows what this looks like for three targets.

\paragraph{The effect follows the manipulation.}
Stability controls also moved least ($\Delta_E=.346$). Displacement exceeded the control baseline for evaluative reframing ($\beta=.101$, [.026, .175]), sense disambiguation ($\beta=.120$, [.028, .212]), and register ($\beta=.147$, [.067, .227]); estimates for event-dependent ($\beta=.074$, [$-.021$, .170]) and mixed affect ($\beta=.081$, [$-.001$, .164]) were less precise. Trial position was unrelated to displacement ($\beta=.015$, [$-.014$, .044]), so fatigue does not explain the effect.

\begin{table}[t]
\centering
\scriptsize
\begin{tabular}{lrrr}
\toprule
Mechanism & Targets & Mean $\Delta_E$ & Coarse switch \\
\midrule
Stability control        & 4  & .346 & 19.9\% \\
Event-dependent affect   & 6  & .427 & 68.8\% \\
Mixed affect             & 8  & .436 & 44.5\% \\
Evaluative reframing     & 17 & .451 & 62.4\% \\
Sense disambiguation     & 13 & .476 & 66.9\% \\
Register                 & 2  & .488 & 55.2\% \\
\bottomrule
\end{tabular}
\caption{Study~2 centroid displacement and coarse-label switch rate by context mechanism. Descriptive: groups differ in size and kind of target.}
\label{tab:study2mechanisms}
\end{table}

\section{Discussion}

Across corpora, valence diagnosticity varies markedly by category, and text-level valence changes with co-occurring labels. The projection evaluation shows that conventional mappings perform well on clearly polar texts, but errors concentrate in opposed-label configurations. These configurations account for 11.8\% of labeled SemEval test tweets, yet 54.8\% of polar prediction errors under learned binary averaging. They are also more than twice as likely to receive neutral-or-mixed V-oc labels. GoEmotions supplies complementary evidence at the within-annotator level: 16.28\% of multilabel emotional judgments combine positive- and negative-parent emotions. The hashtag corpora show only moderate lexical separation between joy and negative emotion signals.

Study~2 shows the same thing one level down. The corpora show that a single text can carry opposed emotions; Study~2 shows that a single word can carry opposed readings until context selects one, and that the selection goes both ways. With word and reader fixed, context switched coarse judgments at nearly three times the control rate, and did so by resolving mixed readings rather than by intensifying them. A collapse can be right on average and still misrepresent a large share of individual instances.

Categorical-to-dimensional mappings are useful and often accurate on average \citep{buechel-hahn-2018-emotion,park-etal-2021-dimensional}. However, averaging emotion vectors into one sentence-level VAD target, as in \citet{el-haj-takanami-2023-unifying}, can obscure co-occurring opposites: joy and sadness may yield a neutral-looking average even when both are annotated. Projection is a modeling choice whose information loss depends on the combination of labels.

\paragraph{Implications for practice.}
Four consequences follow for emotion-label harmonization and evaluation. (i)~Use category-specific evidence. Table~\ref{tab:recommendation} groups categories by observed valence diagnosticity; these are empirical priors rather than guarantees of item-level polarity. (ii)~Account for label combinations. Opposed-label tweets are disproportionately difficult for the tested projection methods, even when their gold polarity is positive or negative. (iii)~Preserve neutral-or-mixed outcomes and report abstention separately. Forced binary projections cannot recover this V-oc class, while abstention trades coverage for fewer predictions. (iv)~Evaluate mappings across languages and datasets rather than assuming that category-level associations transfer unchanged. Where possible, estimate $P(V\mid E_1,\ldots,E_k)$ and retain the uncertainty of the resulting projection.

\begin{table}[t]
\centering
\small
\begin{tabular}{p{0.34\columnwidth} p{0.56\columnwidth}}
\toprule
Handling & Emotion categories \\
\midrule
High-diagnosticity prior & joy, love, sadness, anger, disgust \\
Uncertainty-aware prior & optimism, pessimism, fear, trust \\
No category-only polarity & anticipation, surprise \\
\bottomrule
\end{tabular}
\caption{Empirical guidance based on the minimum held-out diagnosticity $D$ across SemEval languages. High diagnosticity: minimum $D\geq.50$; uncertainty-aware: $.15\leq D<.50$; no category-only polarity: minimum $D<.15$. These descriptive groupings are not independently validated projection rules, and $D$ does not specify polarity direction.}
\label{tab:recommendation}
\end{table}

\section{Conclusion}

Emotion categories provide informative valence priors, but their reliability as text-level polarity predictors depends on the categories and their combinations. Conventional projection performs well on texts with clearly positive or negative gold valence, while opposed-label configurations account for a disproportionate share of prediction errors and frequently receive neutral-or-mixed valence labels. GoEmotions shows that individual annotators also select emotions across sentiment parents, and the controlled annotation study demonstrates that context can substantially change the affective interpretation of the same lexical item without increasing its intensity. Emotion-to-valence mappings should therefore be evaluated at the level of label configurations, with explicit reporting of coverage, uncertainty, and the information lost through coarse projection.

\section*{Limitations}

Study~1 is observational. In SemEval-2018, co-occurrence effects are properties of the annotated tweet: they may reflect distinct clauses, events, or spans and do not show that an emotion itself changes valence when it co-occurs with another. Three languages do not establish broad cross-linguistic generality, and the Affective Text--EmoBank, GoEmotions, and hashtag analyses are English-only.

The projection evaluation starts from gold emotion labels, so it gives an upper bound for label-based projection: a pipeline using predicted labels inherits these errors in addition to its own. The baselines are simple label-only methods and do not establish the best performance achievable from emotion labels; models that also see the text may resolve some opposed configurations. The V-oc classes are derived from continuous V-reg scores rather than elicited independently, and class 0 merges neutral and mixed valence, so we cannot separate texts that annotators found mixed from texts they found neutral. Opposed E-c labels aggregate judgments across annotators and need not mean that any single annotator selected both.

The GoEmotions analysis audits a published hierarchy rather than comparing labels with an independently elicited valence layer, and positive--negative co-selection may partly reflect discourse composition that punctuation-based sentence counts only approximate. Raw judgments are dependent within items and raters, which we address with comment- and rater-cluster bootstraps; consensus-filtered rates are item-level and not directly comparable with rater-level rates. Hashtag labels are self-assigned and noisy, the strict TUSC set is small and dominated by sadness, and lexical valence is a coarse proxy for text valence.

Study~2 used a convenience sample of university students in a course setting, and completion varied: eight main-arm participants completed all 150 trials. We keep incomplete participants only where a valid within-participant pair exists and report participant-equal estimates alongside pair-weighted ones. The isolated target always came first; repeated targets were separated by at least eight trials and trial position showed no association with displacement, but order and prior exposure cannot be ruled out. The 50 targets were chosen to diagnose context sensitivity, not to represent English vocabulary, and the mechanism groups are uneven (two register targets, four controls). Euclidean measures treat the wheel as a metric space, which we take as a property of the instrument rather than a claim about affective coordinates. Annotators did not rank their points, so the strongest association is derived from radial intensity rather than elicited directly.

\section*{Ethical Considerations}
Study~2 participants were university students in a course taught by one of the authors. Before taking part, students were informed that the annotation activity formed part of an active research project and that their annotations could be used as data in a research publication. Research use was voluntary: students could opt out without consequence for their grade by completing an alternative task. The annotation platform collected no names, student numbers, or demographic information linked to the annotations; responses were associated only with non-identifying participant codes used to construct within-participant pairs.

The task involved judgments of individual English words and short invented sentences and contained no personal or sensitive questions, deception, audio or video recording, or physical or psychological intervention. Under \textit{author's institute} Ethics Committee 
Guidelines, research meeting these conditions does not require assessment by the university's Ethics Committee. 
The guidelines additionally permit the use of students from the researcher's own faculty where their status as students is not expected to affect the research results, provided that due consideration is given to pressure on students and protection of personal information; the voluntary alternative-task procedure was used for this purpose.

\paragraph{Data access and licensing.}
We use previously released research datasets and document their sources in the corresponding methods sections. EmoBank is distributed under CC BY-SA 4.0. The NRC VAD Lexicon v2.1 permits noncommercial research and educational use but prohibits redistribution. GoEmotions is available through the Google Research repository; the original Reddit content remains subject to applicable third-party rights. SemEval-2018, Affective Text, TEC, and TUSC/ABCDE are used through their released research resources, subject to their respective access conditions. We do not redistribute raw social-media text or NRC lexicon entries. Analysis code and aggregate results are made available separately.

\section*{Acknowledgments}

This work was supported by JSPS KAKENHI Grant Number 24K21058.

We thank Cheolyeon Byun and Jina Lee for their assistance with the maintenance and deployment of Sentimentator. We also thank the students who volunteered to participate in the annotation study.

Computational resources were provided by CSC -- IT Center for Science, Finland, through the Roihu computing environment.

\bibliography{custom}

\appendix

\section{Analysis details}
\label{app:methods}

\paragraph{Diagnosticity and prototypes (SemEval-2018).}
Because $D$ folds AUC around the chance reference of $.5$, its bootstrap interval is descriptive rather than a test of $D=0$. We use 95\% percentile intervals from 5,000 item-level bootstrap resamples within each language; cross-language summaries average the three language-specific estimates within each replicate. Continuous prototypes are training-set mean V-reg scores. Polarity is evaluated separately using official V-oc classes, with $0$ retained as neutral or mixed. The single-label continuous-valence projection test ($R^2$) compares each score with one prototype rather than an aggregate of several; the test subsets contain 70 English, 125 Arabic, and 247 Spanish single-label tweets.

\paragraph{Polarity projection (SemEval-2018).}
We join V-oc to E-c and V-reg by tweet identifier, language, and original split. Of 2,128 test tweets, 52 have no E-c emotion label, leaving 2,076 for evaluation. Conventional lookup assigns positive polarity to joy, love, optimism, and trust, negative polarity to anger, disgust, fear, sadness, and pessimism, and leaves anticipation and surprise unsigned; ties produce abstention. Learned baselines estimate V-oc class distributions from training data, either averaging per-label distributions or smoothing exact-label-set distributions toward those averages. Binary predictions select a polarity, whereas three-class predictions can also select neutral-or-mixed. Opposed configurations contain at least one conventionally positive and one conventionally negative label. We report coverage and error conditional on prediction, and repeat evaluation on the 1,681 test tweets with polar gold labels. Predictions of class $0$ are not treated as abstentions. These comparisons evaluate the specified label-only baselines, not an upper bound on achievable performance.

For each ordered pair $(e,c)$ and language, we compute
$\Delta V_{e,c}=\overline{V}(e,c)-\overline{V}(e,\neg c)$.
Training pairs require at least five $e+c$ tweets per language and the same sign of $\Delta V$ in English, Arabic, and Spanish; held-out dev+test data assess whether that sign replicates. This is a directional discovery--replication design rather than a multiple-testing procedure; held-out intervals use 5,000 bootstrap resamples.

\paragraph{Projection implementation settings.}
The label-based projection methods do not require neural-network training. Per-emotion and exact-label-set valence distributions are estimated from the SemEval training partitions. Exact-label-set estimates use ten pseudocounts with the averaged per-label distribution as the smoothing prior. This was a fixed reference setting, not the result of hyperparameter optimization. Direction-stability estimates use 5,000 bootstrap resamples with random seed 20261011. Performance is evaluated on the original held-out test partitions without fitting to test labels. Error-rate intervals use the Wilson method. No hyperparameter search was performed for these projection baselines.

\paragraph{Affective Text $\times$ EmoBank (Study~1B).}
We use the released combined EmoBank valence score as primary and repeat the analysis for reader- and writer-perspective valence. Because the original test partition is small for several emotions, the all-item analysis is primary. At $\geq75$ several categories become too sparse, so we do not use that threshold.

\paragraph{GoEmotions (Study~1C).}
The ambiguous parent was left unpolarized by design, so the primary test is positive--negative co-selection; broader any-parent crossings are secondary. For rates over all emotional judgments, a cardinality-preserving permutation null (10,000 permutations) reshuffles labels while preserving judgment cardinality and approximately preserving label marginals. For multilabel rates, a matched null preserves the observed cardinality distribution and samples unique labels by their empirical frequencies; we generate 10,000 complete null datasets and compute one-sided Monte Carlo $p$-values (headline comparison $p=.00010$). Category-level purity uses a category-conditioned version of the same null. Dependence is handled with cluster bootstraps over comments and, separately, over raters (all-judgment rate: comment-cluster CI [3.74, 3.95], rater-cluster CI [3.23, 4.45]), and by removing the 1, 3, 5, and 10 most prolific raters. Sentence counts are estimated from punctuation and only approximately bound discourse complexity. The published aggregation rule (labels chosen by at least two annotators) yields 54,263 items, 38,242 with a non-neutral label; rates after filtering are item-level. For a fine-grained label set $F$ and its deterministic parent projection $C=g(F)$, $H(F\mid C)=H(F)-H(C)$; the eight observed coarse signatures have a ceiling of $\log_2 8=3$ bits.

\paragraph{Study~2.}
Angular and cosine measures of the centroid are secondary diagnostics, because centroid direction is unstable near the origin. The Neutral cutoff for the strongest association is varied from $.08$ to $.12$ as a sensitivity check.

\paragraph{Computational resources.}
The analyses were performed on CPU-based computing environments without neural model training or GPU acceleration. The SemEval polarity-projection audit
was implemented in Python 3.9.25 using the standard library. The analyses involved corpus statistics, bootstrap resampling, permutation-based null models,
and label-based projection. We estimate the total computational budget for the reported analyses at approximately 10 CPU-hours.

\section{Additional results for Studies 1A and 1B}
\label{app:study1ab}

This section provides the analyses underlying the robustness results reported for SemEval-2018 and the Affective Text--EmoBank alignment. We report language-specific diagnosticity estimates (Table~\ref{tab:perlangfull}), co-occurrence results (Tables~\ref{tab:cooc} and~\ref{tab:cooc-full1}--\ref{tab:cooc-full3}), and the EmoBank perspective analysis (Table~\ref{tab:emobankpersp}).

\begin{table}[t]
\centering
\scriptsize
\begin{tabular}{lllr}
\toprule
Base & Co-emotion & $\Delta V$ [95\% CI] & $n_{+c}$ \\
\midrule
Love         & Joy       & +.258 [.226,.292] & 543 \\
Sadness      & Joy       & +.228 [.194,.263] & 86 \\
Anticipation & Joy       & +.225 [.192,.258] & 128 \\
Anger        & Joy       & +.202 [.172,.231] & 86 \\
Optimism     & Sadness   & -.205 [-.241,-.166] & 67 \\
Joy          & Anger     & -.199 [-.229,-.169] & 86 \\
Joy          & Sadness   & -.194 [-.229,-.160] & 86 \\
Anticipation & Anger     & -.159 [-.192,-.125] & 75 \\
\bottomrule
\end{tabular}
\caption{Representative co-occurrence shifts in held-out SemEval-2018 data that replicated the training direction in all three languages and whose pooled 95\% intervals exclude zero. $n_{+c}$: held-out items carrying both labels.}
\label{tab:cooc}
\end{table}

\subsection{Language-specific SemEval-2018 results}

\begin{table*}[t]
\centering
\scriptsize
\begin{tabular}{llrll}
\toprule
Emotion & Lang. & $n_e$ & $D$ [95\% CI] & Mean train $V$ \\
\midrule
Joy & En & 450 & 0.883 [0.850,0.912] & 0.644 \\
Joy & Ar & 292 & 0.896 [0.863,0.924] & 0.738 \\
Joy & Es & 229 & 0.860 [0.819,0.896] & 0.690 \\
Love & En & 167 & 0.658 [0.595,0.716] & 0.706 \\
Love & Ar & 261 & 0.782 [0.732,0.830] & 0.707 \\
Love & Es & 43 & 0.698 [0.593,0.793] & 0.708 \\
Sadness & En & 172 & 0.708 [0.646,0.765] & 0.321 \\
Sadness & Ar & 212 & 0.731 [0.678,0.781] & 0.304 \\
Sadness & Es & 111 & 0.597 [0.510,0.677] & 0.333 \\
Anger & En & 198 & 0.703 [0.646,0.758] & 0.332 \\
Anger & Ar & 227 & 0.637 [0.575,0.696] & 0.319 \\
Anger & Es & 189 & 0.599 [0.529,0.665] & 0.329 \\
Optimism & En & 340 & 0.665 [0.606,0.721] & 0.629 \\
Optimism & Ar & 250 & 0.769 [0.720,0.817] & 0.709 \\
Optimism & Es & 70 & 0.496 [0.388,0.599] & 0.668 \\
Disgust & En & 198 & 0.722 [0.669,0.773] & 0.335 \\
Disgust & Ar & 125 & 0.519 [0.447,0.591] & 0.313 \\
Disgust & Es & 117 & 0.533 [0.446,0.615] & 0.333 \\
Pessimism & En & 71 & 0.559 [0.461,0.653] & 0.313 \\
Pessimism & Ar & 151 & 0.573 [0.504,0.636] & 0.339 \\
Pessimism & Es & 154 & 0.409 [0.329,0.488] & 0.367 \\
Fear & En & 80 & 0.482 [0.399,0.566] & 0.371 \\
Fear & Ar & 105 & 0.532 [0.450,0.609] & 0.318 \\
Fear & Es & 47 & 0.321 [0.175,0.458] & 0.380 \\
Trust & En & 44 & 0.352 [0.221,0.479] & 0.614 \\
Trust & Ar & 53 & 0.326 [0.204,0.439] & 0.631 \\
Trust & Es & 31 & 0.356 [0.199,0.509] & 0.633 \\
Anticipation & En & 93 & 0.038 [0.002,0.139] & 0.504 \\
Anticipation & Ar & 86 & 0.150 [0.045,0.254] & 0.441 \\
Anticipation & Es & 63 & 0.133 [0.018,0.250] & 0.535 \\
Surprise & En & 56 & 0.040 [0.003,0.169] & 0.572 \\
Surprise & Ar & 23 & 0.090 [0.004,0.280] & 0.463 \\
Surprise & Es & 25 & 0.099 [0.007,0.257] & 0.515 \\
\bottomrule
\end{tabular}
\caption{Per-language SemEval-2018 continuous-valence results. $n_e$ denotes test items carrying each emotion; $D$ measures threshold-free valence diagnosticity. Mean train $V$ is the continuous V-reg mean and does not define categorical polarity.}
\label{tab:perlangfull}
\end{table*}



\subsection{Emotion co-occurrence effects}
The main text reports representative co-occurrence effects. 
Tables~\ref{tab:cooc-full1}--\ref{tab:cooc-full3} report the complete held-out screening results and give the complete set of support-eligible ordered pairs carried forward from the training analysis.

\begin{table*}[t]
\centering
\scriptsize
\begin{tabular}{lllrrrr}
\toprule
Base & Co-emotion & Train dir. & Held-out $\Delta V$ [95\% CI] & $n_{+c}$ & Confirmed & All 3 \\
\midrule
Love & Joy & positive & +0.258 [+0.226,+0.292] & 543 & 3/3 & yes \\
Sadness & Joy & positive & +0.228 [+0.194,+0.263] & 86 & 3/3 & yes \\
Anticipation & Joy & positive & +0.225 [+0.192,+0.258] & 128 & 3/3 & yes \\
Trust & Joy & positive & +0.207 [+0.166,+0.249] & 118 & 3/3 & yes \\
Optimism & Sadness & negative & -0.205 [-0.241,-0.166] & 67 & 3/3 & yes \\
Optimism & Joy & positive & +0.205 [+0.182,+0.229] & 751 & 3/3 & yes \\
Anger & Joy & positive & +0.202 [+0.172,+0.231] & 86 & 3/3 & yes \\
Pessimism & Joy & positive & +0.202 [+0.168,+0.235] & 53 & 3/3 & yes \\
Joy & Anger & negative & -0.199 [-0.229,-0.169] & 86 & 3/3 & yes \\
Sadness & Optimism & positive & +0.196 [+0.160,+0.236] & 67 & 3/3 & yes \\
Joy & Sadness & negative & -0.194 [-0.229,-0.160] & 86 & 3/3 & yes \\
Anger & Optimism & positive & +0.190 [+0.134,+0.244] & 37 & 3/3 & yes \\
Joy & Pessimism & negative & -0.188 [-0.222,-0.156] & 53 & 3/3 & yes \\
Optimism & Anger & negative & -0.178 [-0.233,-0.126] & 37 & 3/3 & yes \\
Anticipation & Anger & negative & -0.159 [-0.192,-0.125] & 75 & 3/3 & yes \\
Anticipation & Sadness & negative & -0.148 [-0.185,-0.112] & 54 & 3/3 & yes \\
Anticipation & Optimism & positive & +0.148 [+0.109,+0.187] & 103 & 3/3 & yes \\
Anticipation & Disgust & negative & -0.146 [-0.179,-0.113] & 43 & 3/3 & yes \\
Surprise & Anger & negative & -0.145 [-0.202,-0.088] & 27 & 3/3 & yes \\
Surprise & Disgust & negative & -0.143 [-0.200,-0.088] & 25 & 3/3 & yes \\
Anticipation & Pessimism & negative & -0.138 [-0.171,-0.107] & 47 & 3/3 & yes \\
\bottomrule
\end{tabular}
\caption{Directional co-occurrence discoveries carried from training to held-out evaluation (part 1 of 3). ``Confirmed'': held-out languages matching the training direction over languages with sufficient support. Of 68 support-eligible ordered pairs, 62 met the same-direction training criterion; 46 reproduced that direction in all three held-out languages, and 45 of those had pooled 95\% intervals excluding zero.}
\label{tab:cooc-full1}
\end{table*}

\begin{table*}[t]
\centering
\scriptsize
\begin{tabular}{lllrrrr}
\toprule
Base & Co-emotion & Train dir. & Held-out $\Delta V$ [95\% CI] & $n_{+c}$ & Confirmed & All 3 \\
\midrule
Pessimism & Sadness & negative & -0.131 [-0.161,-0.102] & 240 & 3/3 & yes \\
Fear & Sadness & negative & -0.128 [-0.168,-0.083] & 102 & 3/3 & yes \\
Trust & Optimism & positive & +0.107 [+0.060,+0.153] & 114 & 3/3 & yes \\
Joy & Love & positive & +0.102 [+0.085,+0.120] & 543 & 3/3 & yes \\
Disgust & Surprise & positive & +0.101 [+0.055,+0.147] & 25 & 3/3 & yes \\
Anger & Surprise & positive & +0.099 [+0.052,+0.144] & 27 & 3/3 & yes \\
Fear & Anticipation & positive & +0.092 [+0.046,+0.142] & 54 & 3/3 & yes \\
Fear & Anger & negative & -0.089 [-0.125,-0.054] & 92 & 3/3 & yes \\
Disgust & Anger & negative & -0.088 [-0.117,-0.061] & 482 & 3/3 & yes \\
Optimism & Love & positive & +0.088 [+0.057,+0.122] & 402 & 3/3 & yes \\
Anticipation & Fear & negative & -0.087 [-0.133,-0.037] & 54 & 3/3 & yes \\
Anger & Sadness & negative & -0.084 [-0.105,-0.064] & 285 & 3/3 & yes \\
Fear & Disgust & negative & -0.084 [-0.121,-0.047] & 73 & 3/3 & yes \\
Sadness & Anticipation & positive & +0.082 [+0.050,+0.115] & 54 & 3/3 & yes \\
Disgust & Sadness & negative & -0.080 [-0.104,-0.056] & 220 & 3/3 & yes \\
Anger & Disgust & negative & -0.079 [-0.102,-0.055] & 482 & 3/3 & yes \\
Optimism & Anticipation & negative & -0.069 [-0.107,-0.031] & 103 & 3/3 & yes \\
Joy & Optimism & positive & +0.069 [+0.050,+0.086] & 751 & 3/3 & yes \\
Anger & Anticipation & positive & +0.067 [+0.037,+0.097] & 75 & 3/3 & yes \\
Pessimism & Anger & negative & -0.061 [-0.087,-0.034] & 205 & 3/3 & yes \\
Sadness & Anger & negative & -0.060 [-0.082,-0.038] & 285 & 3/3 & yes \\
\bottomrule
\end{tabular}
\caption{Directional co-occurrence discoveries (part 2 of 3).}
\label{tab:cooc-full2}
\end{table*}

\begin{table*}[t]
\centering
\scriptsize
\begin{tabular}{lllrrrr}
\toprule
Base & Co-emotion & Train dir. & Held-out $\Delta V$ [95\% CI] & $n_{+c}$ & Confirmed & All 3 \\
\midrule
Pessimism & Anticipation & positive & +0.057 [+0.026,+0.088] & 47 & 3/3 & yes \\
Anger & Fear & negative & -0.038 [-0.069,-0.008] & 92 & 3/3 & yes \\
Sadness & Pessimism & negative & -0.034 [-0.057,-0.010] & 240 & 3/3 & yes \\
Optimism & Trust & negative & -0.012 [-0.043,+0.020] & 114 & 3/3 & yes \\
Love & Sadness & negative & -0.270 -- & 39 & 2/2 & no \\
Fear & Optimism & positive & +0.209 -- & 24 & 2/2 & no \\
Optimism & Pessimism & negative & -0.200 -- & 20 & 2/2 & no \\
Surprise & Sadness & negative & -0.200 -- & 18 & 2/2 & no \\
Sadness & Love & positive & +0.175 -- & 39 & 2/2 & no \\
Optimism & Fear & negative & -0.158 -- & 24 & 2/2 & no \\
Pessimism & Optimism & positive & +0.138 -- & 20 & 2/2 & no \\
Trust & Love & positive & +0.137 -- & 66 & 2/2 & no \\
Sadness & Surprise & positive & +0.091 -- & 18 & 2/2 & no \\
Disgust & Anticipation & positive & +0.062 [+0.033,+0.093] & 43 & 2/3 & no \\
Sadness & Fear & negative & -0.047 [-0.084,-0.005] & 102 & 2/3 & no \\
Love & Optimism & positive & +0.035 [+0.002,+0.070] & 402 & 2/3 & no \\
Joy & Anticipation & negative & -0.034 [-0.064,-0.002] & 128 & 2/3 & no \\
Disgust & Fear & negative & -0.033 [-0.065,-0.000] & 73 & 2/3 & no \\
Pessimism & Disgust & negative & -0.028 [-0.057,+0.000] & 171 & 2/3 & no \\
Pessimism & Fear & negative & -0.015 [-0.048,+0.020] & 67 & 2/3 & no \\
\bottomrule
\end{tabular}
\caption{Directional co-occurrence discoveries (part 3 of 3).}
\label{tab:cooc-full3}
\end{table*}

\subsection{EmoBank perspective robustness}
The main analysis uses combined EmoBank valence. Results are similar when reader and writer perspectives are analyzed separately.

\begin{table*}[t]
\centering
\scriptsize
\begin{tabular}{lrlllrrr}
\toprule
Emotion & $n$ & Reader $D$ [95\% CI] & Writer $D$ [95\% CI] & Combined $D$ [95\% CI] &
Reader $\rho$ & Writer $\rho$ & Combined $\rho$ \\
\midrule
Anger    & 38  & .614 [.502,.723] & .472 [.328,.609] & .603 [.493,.706] & -.545 & -.479 & -.558 \\
Disgust  & 19  & .642 [.537,.742] & .562 [.374,.725] & .661 [.539,.769] & -.404 & -.346 & -.407 \\
Fear     & 114 & .614 [.534,.688] & .467 [.379,.549] & .609 [.535,.678] & -.549 & -.474 & -.562 \\
Joy      & 144 & .767 [.720,.810] & .759 [.707,.810] & .843 [.808,.875] & +.731 & +.684 & +.773 \\
Sadness  & 139 & .793 [.747,.836] & .752 [.692,.808] & .832 [.787,.873] & -.677 & -.632 & -.709 \\
Surprise & 49  & .217 [.061,.370] & .171 [.027,.320] & .220 [.065,.370] & +.123 & +.142 & +.143 \\
\bottomrule
\end{tabular}
\caption{Affective Text $\times$ EmoBank across reader, writer, and combined valence. The ordering is preserved in all three; the combined score is slightly more diagnostic for joy and sadness, as expected from averaging two noisy ratings.}
\label{tab:emobankpersp}
\end{table*}

\section{Additional analyses for Study 1C}
\label{app:goemotions}

The GoEmotions analyses in the main text aggregate across categories. Table~\ref{tab:goemotioncats} reports category-level hierarchy conformity, and Table~\ref{tab:raterdrop} tests sensitivity to the most prolific annotators.

\subsection{Category-level hierarchy results}
\begin{table*}[t]
\centering
\scriptsize
\begin{tabular}{llrrrrr}
\toprule
Emotion & Parent & $n_{ML}$ & Purity obs. & Purity null & Opp./val. obs. & Opp./val. null \\
\midrule
Admiration & positive & 6600 & 0.721 & 0.407 & 0.130 & 0.418 \\
Amusement & positive & 3115 & 0.574 & 0.436 & 0.220 & 0.387 \\
Approval & positive & 6361 & 0.617 & 0.412 & 0.167 & 0.410 \\
Caring & positive & 2476 & 0.628 & 0.431 & 0.251 & 0.399 \\
Desire & positive & 1670 & 0.620 & 0.432 & 0.213 & 0.399 \\
Excitement & positive & 2609 & 0.729 & 0.425 & 0.086 & 0.406 \\
Gratitude & positive & 4550 & 0.702 & 0.427 & 0.155 & 0.398 \\
Joy & positive & 3654 & 0.799 & 0.428 & 0.088 & 0.400 \\
Love & positive & 3234 & 0.736 & 0.434 & 0.163 & 0.395 \\
Optimism & positive & 4196 & 0.649 & 0.422 & 0.209 & 0.405 \\
Pride & positive & 612 & 0.778 & 0.387 & 0.132 & 0.449 \\
Relief & positive & 501 & 0.683 & 0.439 & 0.216 & 0.395 \\
Anger & negative & 2882 & 0.706 & 0.252 & 0.186 & 0.587 \\
Annoyance & negative & 5276 & 0.657 & 0.245 & 0.214 & 0.590 \\
Disappointment & negative & 3763 & 0.604 & 0.244 & 0.260 & 0.601 \\
Disapproval & negative & 3738 & 0.565 & 0.256 & 0.284 & 0.584 \\
Disgust & negative & 2387 & 0.650 & 0.251 & 0.227 & 0.596 \\
Embarrassment & negative & 1043 & 0.583 & 0.240 & 0.264 & 0.618 \\
Fear & negative & 1419 & 0.607 & 0.274 & 0.256 & 0.575 \\
Grief & negative & 322 & 0.624 & 0.256 & 0.280 & 0.598 \\
Nervousness & negative & 1014 & 0.669 & 0.259 & 0.215 & 0.594 \\
Remorse & negative & 1015 & 0.464 & 0.265 & 0.377 & 0.583 \\
Sadness & negative & 2931 & 0.531 & 0.249 & 0.330 & 0.594 \\
Confusion & ambiguous & 2421 & 0.363 & 0.114 & 0.637 & 0.886 \\
Curiosity & ambiguous & 3807 & 0.308 & 0.097 & 0.692 & 0.903 \\
Realization & ambiguous & 4071 & 0.165 & 0.097 & 0.835 & 0.903 \\
Surprise & ambiguous & 2042 & 0.286 & 0.114 & 0.714 & 0.886 \\
\bottomrule
\end{tabular}
\caption{GoEmotions category-level multilabel results. For positive and negative categories, ``Opp./val.'' is opposite-valence co-selection; for ambiguous categories, co-selection with either valenced parent. Nulls preserve the cardinality distribution of judgments containing each category and sample remaining labels by empirical frequency.}
\label{tab:goemotioncats}
\end{table*}

\subsection{Rater sensitivity}
\begin{table}[t]
\centering
\scriptsize
\begin{tabular}{lrr}
\toprule
Rater set & Positive+negative & Any-parent \\
\midrule
All raters & 3.84\% & 9.30\% \\
Drop top 1 & 3.93\% & 9.47\% \\
Drop top 3 & 3.62\% & 7.98\% \\
Drop top 5 & 3.51\% & 7.49\% \\
Drop top 10 & 3.67\% & 8.13\% \\
\bottomrule
\end{tabular}
\caption{GoEmotions sensitivity to prolific raters, over all emotional rater-level judgments.}
\label{tab:raterdrop}
\end{table}

\section{Additional analyses for Study 1D}
\label{app:hashtags}

This section reports descriptive lexical-valence distributions and the expanded-hashtag robustness analysis for the naturalistic corpora.

\begin{table}[t]
\centering
\tiny
\begin{tabular}{llrll}
\toprule
Corpus & Joy vs. & $n$ & AUC [95\% CI] & $D$ [95\% CI] \\
\midrule
TEC  & pooled negative & 17,144 & .666 [.658,.675] & .333 [.317,.349] \\
     & sadness         & 12,036 & .648 [.638,.658] & .296 [.275,.316] \\
     & fear            & 11,024 & .639 [.627,.651] & .278 [.254,.303] \\
     & anger           &  9,778 & .736 [.723,.749] & .472 [.445,.498] \\
     & disgust         &  8,987 & .716 [.698,.735] & .433 [.395,.469] \\
\midrule
TUSC & pooled neg., strict   & 332 & .716 [.654,.776] & .433 [.308,.552] \\
     & pooled neg., expanded & 946 & .808 [.779,.836] & .616 [.558,.672] \\
\bottomrule
\end{tabular}
\caption{Study~1D: how well residual lexical valence separates hashtag-labeled joy from negative categories after all hashtags are removed. $n$: tweets in the comparison. 95\% intervals from 5,000 bootstrap resamples.}
\label{tab:tec}
\end{table}

\subsection{TEC lexical-valence distributions}
\begin{table}[t]
\centering
\scriptsize
\begin{tabular}{lrrr}
\toprule
Emotion & $n$ & Mean lexical $V$ & Median lexical $V$ \\
\midrule
Joy     & 8,227 & .204 & .199 \\
Sadness & 3,809 & .120 & .125 \\
Fear    & 2,797 & .116 & .123 \\
Disgust &   760 & .077 & .089 \\
Anger   & 1,551 & .066 & .077 \\
\bottomrule
\end{tabular}
\caption{Mean lexical valence in TEC by hashtag emotion after removal of hashtag, URL, and mention material.}
\label{tab:tecmeans}
\end{table}

\subsection{TUSC/ABCDE hashtag robustness}
The ABCDE/TUSC strict inventory uses direct affective variants: \texttt{joy/joyful/happy/happiness}, \texttt{sad/sadness}, \texttt{fear/fearful/scared}, \texttt{anger/angry}, and \texttt{disgust/disgusted}. The expanded inventory adds broader affective synonyms and hashtag phrases; the complete seed list is provided with the analysis code. All hashtags are removed before lexical scoring.

\begin{table}[t]
\centering
\scriptsize
\begin{tabular}{lrrrr}
\toprule
Inventory & $n$ & AUC [95\% CI] & $D$ [95\% CI] & Opp.-only$^\dagger$ \\
\midrule
Strict   & 332 & .716 [.654,.776] & .433 [.308,.552] & 26.8\% \\
Expanded & 946 & .808 [.779,.836] & .616 [.558,.672] & 12.9\% \\
\bottomrule
\end{tabular}
\caption{ABCDE/TUSC joy-versus-pooled-negative robustness analysis after removal of all hashtags. $^\dagger$Among posts containing at least one extreme lexical cue ($|V|\geq.70$), the percentage containing extreme cues exclusively opposite to the hashtag-derived valence. The strict and expanded analyses contain 231 and 728 extreme-cued posts, respectively.}
\label{tab:abcde}
\end{table}

\section{Study 2 materials and supplementary analyses}
\label{app:study2}

This section documents the lexical material used to construct Study~2
and reproduces the annotation instructions shown to participants.

\subsection{Candidate-generation data}

\begin{table*}[h!]
\centering
\scriptsize
\begin{tabular}{lrrrrr}
\toprule
Lexical item & Positive (\%) & Negative (\%) & Neither (\%) & Both (\%) & $H_{\mathrm{norm}}$ \\
\midrule
bear       & 8.7  & 34.8 & 26.1 & 30.4 & .932 \\
escape     & 26.1 & 21.7 & 8.7  & 43.5 & .907 \\
happen     & 4.3  & 17.4 & 30.4 & 47.8 & .833 \\
stare      & 0.0  & 47.8 & 34.8 & 17.4 & .739 \\
fire       & 0.0  & 21.7 & 21.7 & 56.5 & .711 \\
surprise   & 34.8 & 0.0  & 8.7  & 56.5 & .651 \\
morning    & 65.2 & 0.0  & 13.0 & 21.7 & .632 \\
help       & 69.6 & 13.0 & 0.0  & 17.4 & .593 \\
time       & 4.5  & 0.0  & 59.1 & 36.4 & .591 \\
continue   & 69.6 & 0.0  & 21.7 & 8.7  & .575 \\
strike     & 0.0  & 72.7 & 13.6 & 13.6 & .559 \\
promise    & 78.3 & 0.0  & 8.7  & 13.0 & .483 \\
hide       & 0.0  & 78.3 & 17.4 & 4.3  & .456 \\
cold       & 0.0  & 81.8 & 9.1  & 9.1  & .433 \\
fall       & 0.0  & 81.8 & 13.6 & 4.5  & .416 \\
cry        & 0.0  & 78.3 & 0.0  & 21.7 & .378 \\
brother    & 78.3 & 0.0  & 21.7 & 0.0  & .378 \\
strong     & 87.0 & 0.0  & 4.3  & 8.7  & .339 \\
home       & 82.6 & 0.0  & 17.4 & 0.0  & .333 \\
strange    & 0.0  & 87.0 & 13.0 & 0.0  & .279 \\
doubt      & 0.0  & 91.3 & 4.3  & 4.3  & .257 \\
suspect    & 0.0  & 91.3 & 4.3  & 4.3  & .257 \\
lose       & 0.0  & 91.3 & 0.0  & 8.7  & .213 \\
laugh      & 91.3 & 0.0  & 0.0  & 8.7  & .213 \\
die        & 0.0  & 91.3 & 0.0  & 8.7  & .213 \\
hell       & 0.0  & 91.3 & 8.7  & 0.0  & .213 \\
pretty     & 95.7 & 0.0  & 0.0  & 4.3  & .129 \\
together   & 95.7 & 0.0  & 0.0  & 4.3  & .129 \\
lie        & 0.0  & 95.7 & 0.0  & 4.3  & .129 \\
dark       & 0.0  & 95.7 & 0.0  & 4.3  & .129 \\
light      & 95.7 & 0.0  & 4.3  & 0.0  & .129 \\
hope       & 100.0 & 0.0   & 0.0 & 0.0 & .000 \\
bad        & 0.0   & 100.0 & 0.0 & 0.0 & .000 \\
kill       & 0.0   & 100.0 & 0.0 & 0.0 & .000 \\
grow       & 100.0 & 0.0   & 0.0 & 0.0 & .000 \\
party      & 100.0 & 0.0   & 0.0 & 0.0 & .000 \\
friend     & 100.0 & 0.0   & 0.0 & 0.0 & .000 \\
depressed  & 0.0   & 100.0 & 0.0 & 0.0 & .000 \\
struggle   & 0.0   & 100.0 & 0.0 & 0.0 & .000 \\
murder     & 0.0   & 100.0 & 0.0 & 0.0 & .000 \\
steal      & 0.0   & 100.0 & 0.0 & 0.0 & .000 \\
\bottomrule
\end{tabular}
\caption{Earlier lexical annotation data used as one source for Study~2 candidate generation. Twenty-three annotators judged 41 isolated English words as Positive, Negative, Neither, or Both; four of 943 possible judgments are missing. $H_{\mathrm{norm}}$ is normalized four-category response entropy. Items are ordered from highest to lowest disagreement.}
\label{tab:legacy-lexical}
\end{table*}

\subsection{Annotation guidelines}
\begin{figure*}[h!]
    \centering
    \includegraphics[width=\linewidth]{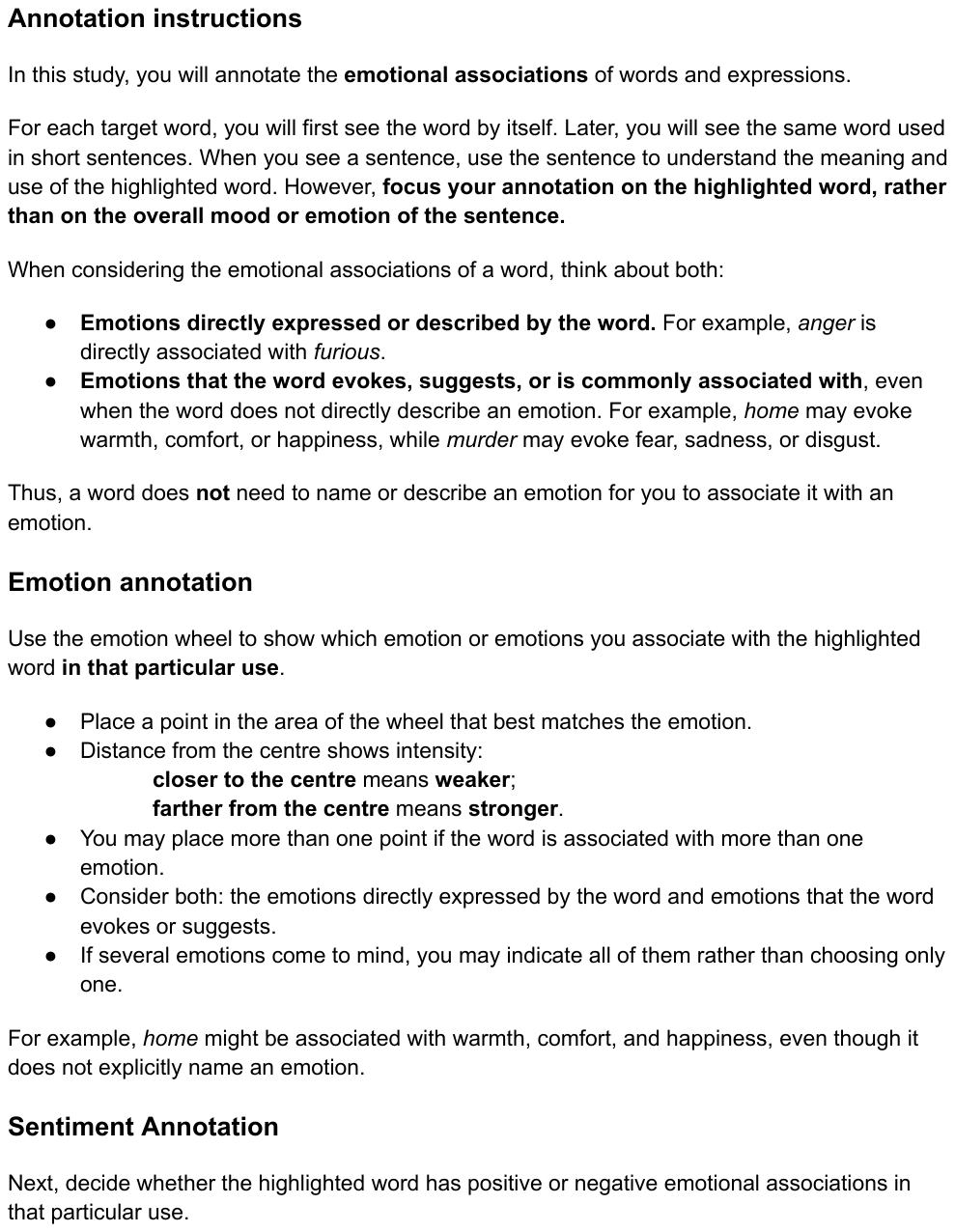}
\end{figure*}

\subsection{Annotation guidelines}
\begin{figure*}[h!]
    \centering
    \includegraphics[width=\linewidth]{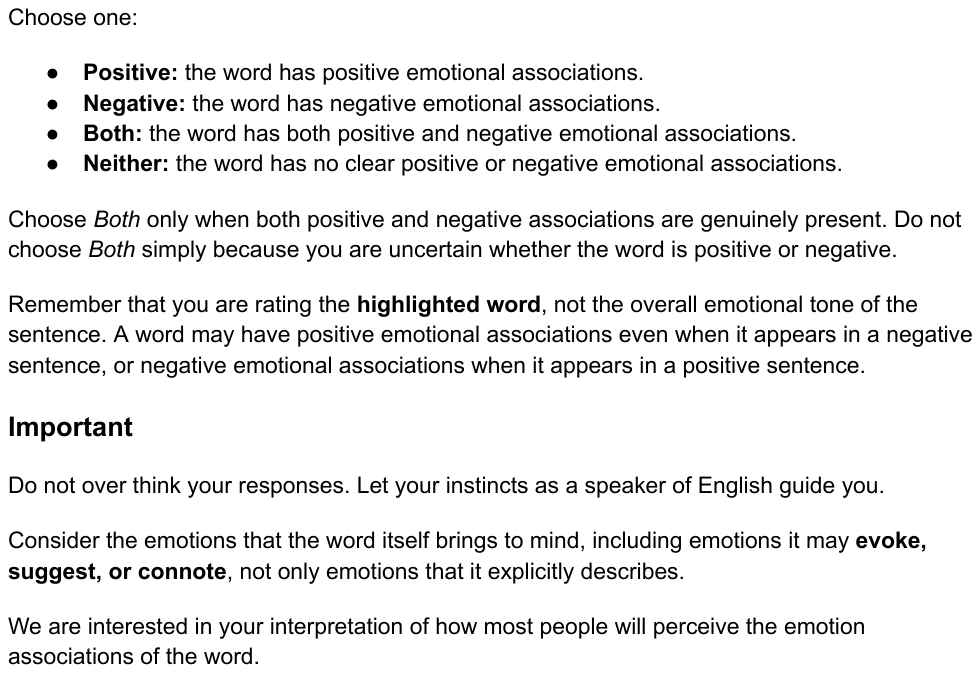}
\end{figure*}

\section{Full Transition Matrix}

\begin{figure}[h!]
\centering
\includegraphics[width=\columnwidth]{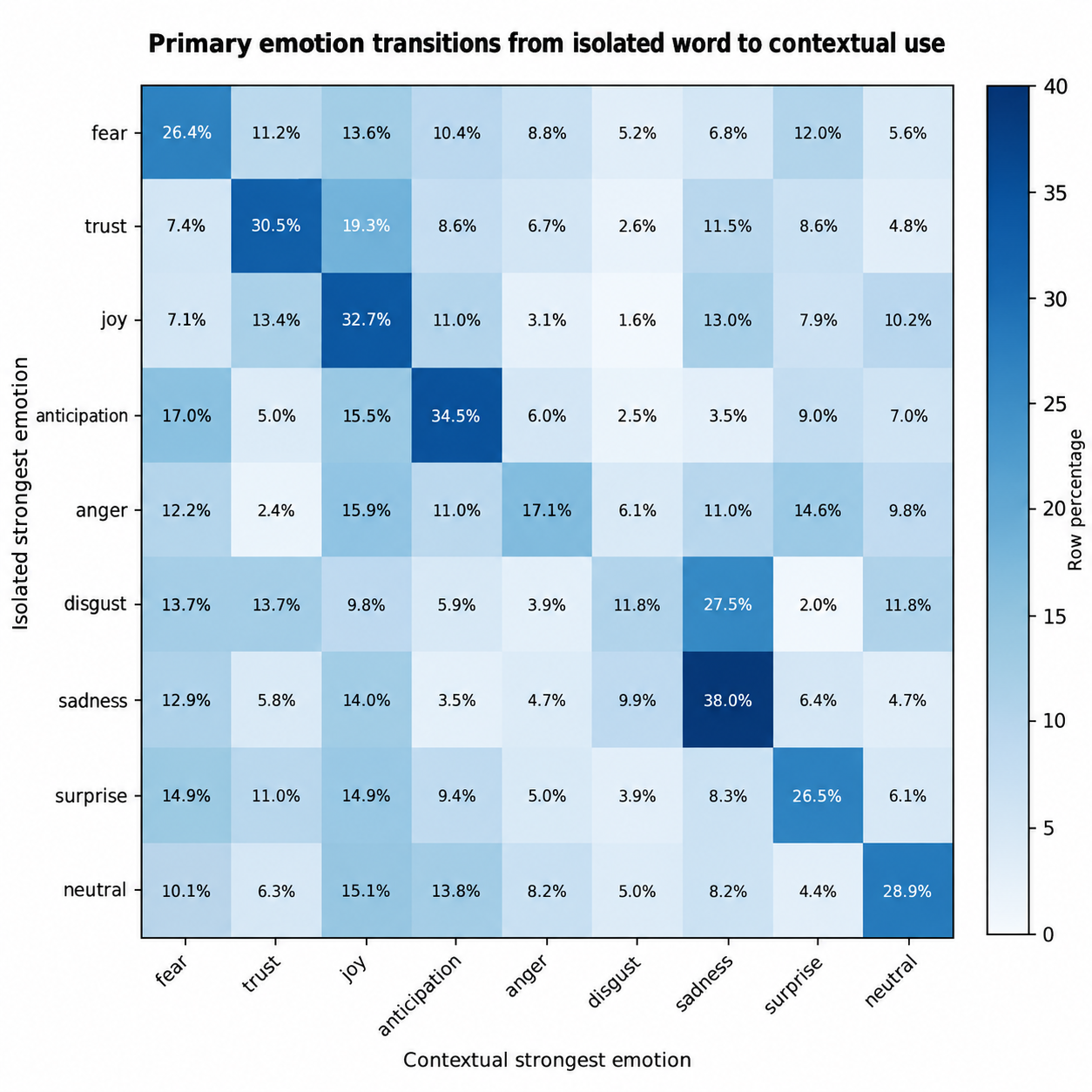}
\caption{Row-normalized transition matrix for the strongest emotion association from the isolated lexical target to a contextualized use. Rows are the isolated strongest-emotion sector and columns the contextual strongest-emotion sector.}
\label{fig:study2transitions}
\end{figure}

\section{Empirical guide to emotion-to-valence projection}
\label{app:projectionguide}

Table~\ref{tab:projectionguide} reports the minimum held-out diagnosticity $D$ across SemEval languages, so weakness in one language is not hidden by averaging. The handling categories are descriptive recommendations based on diagnosticity rather than validated polarity decision rules. Continuous-valence diagnosticity does not establish the direction or certainty of ordinal polarity.

\begin{table*}[t]
\centering
\scriptsize
\begin{tabular}{lrll}
\toprule
Emotion & Min. $D$ & Study~1B & Handling \\
\midrule
Joy          & 0.860 & strong       & high-diagnosticity prior \\
Love         & 0.658 & --           & high-diagnosticity prior \\
Sadness      & 0.597 & strong       & high-diagnosticity prior \\
Anger        & 0.599 & intermediate & high-diagnosticity prior \\
Disgust      & 0.519 & intermediate & high-diagnosticity prior \\
Optimism     & 0.496 & --           & uncertainty-aware prior \\
Pessimism    & 0.409 & --           & uncertainty-aware prior \\
Fear         & 0.321 & intermediate & uncertainty-aware prior \\
Trust        & 0.326 & --           & uncertainty-aware prior \\
Anticipation & 0.038 & --           & no category-only polarity \\
Surprise     & 0.040 & weak         & no category-only polarity \\
\bottomrule
\end{tabular}
\caption{Reference diagnosticity values and empirical handling recommendations. Study~1B gives cross-dataset evidence for six categories. Groupings follow Table~\ref{tab:recommendation} and are not validated prediction rules.}
\label{tab:projectionguide}
\end{table*}

\paragraph{Multilabel projection.}
Table~\ref{tab:projectionguide} describes category-level evidence rather than item-level inheritance rules. Opposed emotion labels indicate a configuration for which the tested polarity projections are less reliable; they do not establish that a text is itself mixed. Where possible, project the complete label configuration using an estimated $P(V\mid E_1,\ldots,E_k)$, retain uncertainty, and evaluate neutral-or-mixed outcomes separately. In GoEmotions, 33.0\% of multilabel judgments containing sadness also contain a positive-parent emotion, illustrating information that would be discarded by an exclusive single-polarity representation.

\end{document}